\documentclass[10pt,compsoc,conference]{IEEEtran}
\IEEEoverridecommandlockouts
\usepackage[utf8]{inputenc}
\usepackage{amsmath,amssymb,amsfonts}
\usepackage[linesnumbered,ruled,vlined]{algorithm2e}
\usepackage{graphicx}
\usepackage{xspace}
\usepackage{booktabs}
\usepackage{subfig}
\usepackage{fnpct}
\usepackage{pgfplots}
\pgfplotsset{compat=1.14}

\usepackage{comment}

\usepackage[backend=biber,bibstyle=ieee,citestyle=numeric]{biblatex}
\bibliography{ms.bib}

\newtheorem{definition}{Definition}

\newcommand{\T}{\ensuremath{\mathcal{T}}\xspace}

\begin{document}

\title{Towards Testing of Deep Learning Systems with Training Set Reduction}

\author{\IEEEauthorblockN{Helge Spieker}
\IEEEauthorblockA{\textit{Simula Research Laboratory}\\
Lysaker, Norway \\
helge@simula.no}
\and
\IEEEauthorblockN{Arnaud Gotlieb}
\IEEEauthorblockA{\textit{Simula Research Laboratory}\\
Lysaker, Norway \\
arnaud@simula.no}
\thanks{This work is supported by the Research Council of Norway (RCN) through the research-based innovation center Certus, under the SFI program.}}

\maketitle

\begin{abstract}
Testing the implementation of deep learning systems and their training routines is crucial to maintain a reliable code base.
Modern software development employs processes, such as Continuous Integration, in which changes to the software are frequently integrated and tested.
However, testing the training routines requires running them and fully training a deep learning model can be resource-intensive, when using the full data set. 
Using only a subset of the training data can improve test run time, but can also reduce its effectiveness.
We evaluate different ways for training set reduction and their ability to mimic the characteristics of model training with the original full data set.
Our results underline the usefulness of training set reduction, especially in resource-constrained environments.
\end{abstract}

\begin{IEEEkeywords}
Testing of Machine Learning, Deep Learning, Continuous Integration, Software Testing
\end{IEEEkeywords}

\section{Introduction}

This paper proposes a method to reduce the training set of a deep learning (DL) model for testing purposes.
The reduced training set allows faster model training than the original, full-sized training set and mimics the training characteristics of full-sized model training.

The increasing number of software systems based on machine learning systems and learning components requires adequate testing techniques to ensure their functionality.
Continuous Integration (CI) is a software development technique for iterative software development. 
Within CI cycles, changes to the software are integrated into the main software repository, built and tested.
One major aspect of modern CI systems is to focus on short cycles times, that provide feedback about the change quality early after submitting the changes to version control.

As deep learning (DL), as well as other machine learning systems are more prevalent and central to the software success, also their development and quality assurance is moved within the CI paradigm. 
However, to extensively test the functionality of DL training and optimization, it is often necessary to perform training runs of the systems, which is a resource-intensive task, both regarding time and computing resources. 
Therefore, there is a need to develop techniques to find alternatives to full-sized model training for resource-saving and reduced feedback times.

With training set reduction (TSR), this issues is mitigated by reducing the number of training samples and thereby the computational requirements for model training.
While this reduced training set does not yield the same model quality, it aims to mimic training behaviours of the original full-sized training process. 
Conclusively, only using a subset of the data can allow to improve the testing process of DL systems.
In this paper, we discuss and evaluate three approaches to training set reduction, namely distance-based reduction to cover the input space, loss-based reduction to focus on the most difficult instances, and random selection, which is most widely used in practice, e.g. in \cite{Dwarakanath2018}.

The goal is to have a reduced training set whose behaviour in terms of loss development best mimics the full training set.
The motivation for this is to be able to allow to test for differences in model training and varying loss curves.
We test seven different training set sizes for each strategy on two deep neural network architectures, using the CIFAR-10 data set \cite{Krizhevsky2014cifar}.

While the focus of this paper lies on DL systems, including experimental evaluation, the methods are similarly transferable to other data-driven ML systems.
Nevertheless, we choose DL systems as a first application method due to all the wide variety of different machine learning methods, that is difficult to cover, the recent successes and widespread usages of DL systems, that also highlight an increasing necessity for testing methods, and the often especially resource-intensive training processes for DL systems.

\section{Background}

\subsection{Training Neural Networks} \label{sec:nntraining}

The effectiveness of artificial neural networks, especially deep neural networks (DNNs), lies, among algorithmic and hardware developments, in the availability of large amounts of data for training.
In contrast to traditional software, neural networks are defined and programmed by their network architecture, their training data and a stochastic training method.
During repeated training epochs, i.e. iterations in which the whole data set is seen by the network, the network learns to identify features and patterns in the training data and to perform the desired task, e.g. classifying the image into one of multiple categories.

A neural network is trained from a training set, consisting of example instances $(x_i,y_i)$.
Training the model describes the process to minimize the error between estimated value $\hat{y}$ and true (observed) value $y$ of the training examples.
The error is assessed with a loss function, that can be different depending on the task of the machine learning model.
For regression problems, where a continuous output value is estimated, the loss is commonly assessed via the \emph{mean squared error}.
In classification, where the input is assigned to one of multiple classes, the \emph{cross-entropy loss} is calculated.

\begin{definition}[Mean Squared Error (MSE)]
Given a set of $N$ estimated and observed target values $\{(\hat{y}_1, y_1),(\hat{y}_2, y_2),\dots,(\hat{y}_N, y_N)\}$,
the MSE is calculated as: 
\[L = \frac{1}{N} \sum_{i=1}^{N} (\hat{y}_i - y_i)^2\]

Within the calculation of MSE, positive and negative errors have the same effect, but larger errors are stronger penalized, i.e. they have a larger influence, than smaller errors.
\end{definition}

\begin{definition}[Cross-Entropy Loss]
Given a set of $N$ estimates and $K$ classes, the cross-entropy (also log-likelihood) is calculated as:
\[L = -\frac{1}{N} \sum_{i=1}^{N} \sum_{k=1}^{K} y_{ik} \log \hat{y_{ik}}\]
with $\hat{y_{ik}}$ being the probability of $x_i$ belonging to class $k$ and 
\[
y_{ik} =
\begin{cases}
1 & \text{if $x_i$ belongs to class $k$}\\
0 & \text{otherwise}
\end{cases}
\]
\end{definition}

In addition to the training set, a second, independent set of data is used to frequently evaluate the performance of the model.
The information from this validation set is only used to monitor the model performance, but the samples in this data set are not used to adjust the weights.

\subsection{Neural Network Architectures} \label{sec:architectures}

A neural network is organized in a number of layers, each consisting of a number of hidden nodes.
Each node is associated with a weight, which is adjusted during the training of the network.
Data is passed through the network layer-wise from the initial input layer through each of the intermediate hidden layers to the final output layer.
At each layer, the outputs of the previous layer are modified by multiplication with a node's weight and applying a non-linear activation function.
Popular activation functions include the sigmoid function, tanh (hyperbolic tangent), and ReLU (Rectified Linear Unit, filtering values below 0).
The final output layer consolidates the outputs and either predicts a continuous output value, in case of a regression task, or predicts a probability for each of a number of classes, in case of a classification task.

Of particular importance, especially for image analysis tasks, are neural networks with convolutional layers.
These layers allow to find similar spatial patterns regardless of their explicit position within the input image \cite{Lecun2010}.
Most current neural network architectures for image analysis include convolutional layers within their building blocks.

We will first describe a simple CNN architecture to which we will further refer as \emph{SimpleCNN}, that can be applied to basic image classification tasks. 
The network consists of two convolutional layers, which apply a number of filters on the input to build a spatial feature map.
Each of these layers is followed by a max-pool layer, that groups the features for neighbouring pixels to reduce dimensionality and highlight relevant parts of the image.
This initial convolutional part is designed to learn and select important features within the input images.
Afterwards, the result is passed to two fully-connected output layer, that predict to which class the image belongs.
This model architecture is certainly to simple and small for large-scale image classification tasks, but it is less complex to understand.
As this architecture is also used in image classification tutorials\footnote{\url{https://pytorch.org/tutorials/beginner/blitz/cifar10_tutorial.html}}, it is often a starting for the development of new models on both known and novel data sets.

The second DNN architecture, that will be further used throughout this paper, is a residual neural network (ResNet) \cite{He2015,He2016}.
ResNets introduced a new building block for DNN architectures, the residual block.
In a residual block, which can be loosely described as a more complex neural network layer, the layer output consists of both the activation functions outcomes plus the layer input.
The task of each layer is thereby only to learn the difference to be applied to the layer input, i.e. the residual between layers.
Applying these residual skip connections showed to be highly effective and allowed the authors to train very deep networks with up to 150 layers, but also increased the performance of ResNet models with fewer layers.
Common sizes for practical applications are 18, 34 or 50 layers, as defined in \cite{He2015}.

We refer the interested reader to \cite{James2013,Bengio2012a,Domingos2012} for discussions of practical considerations when defining neural network architectures and their training process. 

\subsection{Related Work} \label{sec:relwork}

The usage ML and DL for testing traditional software systems has been successfully investigated in a variety of testing aspects, such as compiler fuzzing \cite{Cummins2018}, test case prioritization \cite{Spieker2017,Chen2017}, test case generation \cite{Reichstaller2018}, or failure analysis \cite{Rosenberg2018}.
However, recently, also the testing of DL systems has received increasing attention from both the software testing community as well as DL researchers.
Proposed techniques include differential and multi-implementation testing \cite{Pei2017,Alebiosu2018}, mutation testing \cite{Ma2018}, and metamorphic testing \cite{Murphy2008,Dwarakanath2018}.
To evaluate test adequacy, in relation to the code coverage criterion used in traditional software testing, neuron coverage \cite{Pei2017} and layer coverage \cite{Ma2018} criteria have been proposed.
These criteria measure to which degree the individual nodes of a network have been activated by a set of test inputs.
Nevertheless, as in traditional software testing, maximizing these coverage criteria does only serve as an indicator for test quality, but does not guarantee failure-absence.
In relation to this criteria, traditional software testing applies mutation testing techniques to evaluate an existing test suite's quality \cite{Jia2011,Papadakis2018b} and a similar approach has been proposed for testing DL systems \cite{Ma2018}.

The effects of training set sizes on ML systems have been extensively studied in the past, especially during early developments of neural network techniques in the 1990s \cite{Foody1995,Chen1996}, and in the context of instance-based algorithms \cite{Chen1996,Wilson2000,Sanchez2004}, such as k-nearest neighbour methods, which rely on the training set being available at prediction time and were a reduced data set size results in runtime improvements.

\section{Training Set Reduction}

\subsection{Overview}

Training set reduction (TSR) is a parameterized technique, that receives an initial, full-sized data set \T plus a size parameter $s$ as an input and returns a reduced data set $\T_{red}$ of the given size.
The optimization goal of TSR is to select \T such that the training process of a ML model shows similar properties $P$:
\begin{align*}
\text{Maximize }& similarity(P(\T), P(\T_{red}))\\
\text{with }& \T_{red} = \text{TSR}(\T, s)\\
\text{and }& \T_{red} \subsetneq \T
\end{align*}

Two parameters in the problem definition are externally given, the similarity function and the size of the reduced training set $s$.
The training set size $s$ is a parameter to adjust the trade-off between improvements in runtime and similarity to the original properties.
The choice of $s$ is therefore in conflict with the optimization objective, as a larger subset of data can better approximate the whole data set.
The similarity function is a non-specific comparison function to compare the characteristics of two training runs or, more general, two sets of training runs.

For the remainder of this paper, we choose $similarity$ such that it compares the curvature for training and validation loss curves, but other similarity functions are possible and can fit into the TSR problem description.
However, it is to be noted, that for the TSR strategies presented next, the similarity function does not have to be present in order to derive $T_{red}$.
An exhaustive or guided search under consideration of the similarity as an optimization objective would require frequent re-training of the ML model, which is costly.
The same applies to the automatic selection of the subset size $s$.
We therefore focus on the design of heuristic strategies for TSR.
Under this consideration, the similarity function mostly has the purpose to evaluate the decisions made by each strategy.

With all TSR strategies, it is important to maintain a resemblance of the full-sized data set in both the input and the output space.
For classification tasks, that is, when an input sample has to be assigned to one of a predefined number of classes, the distribution of classes in the reduced data set should be almost identical to the original data set.
Alternatively, for regression tasks, where a continuous value is predicted by the model, the output values of the original data set can be discretized and the distribution can then be preserved similarly to a classification task.
From this distribution, it can then be calculated how many samples for class or discrete value should remain in the resulting subset.
Through this approach, the distribution of output values under the reduced data set remains similar to the distribution under the full-sized data.

\subsection{Subset Selection Strategies} \label{sec:strategies}

In the following, we discuss three TSR strategies.
Two systematic TSR strategies consider additional information within the full-sized data set to select the most representative samples.
The third TSR strategy, random subset selection, does not consider any additional information, but chooses the subset data set by uniform sampling.

\subsubsection{Distance-based Subset Selection}
This selection strategy aims to cover the original data set by selecting instances based on their distance to each other.
It relies on having a clear sample distinction, either by predefined classes in a classification task, or discretization in a regression task, as described before.

If the requirements are fulfilled, the strategy proceeds by clustering the input samples for each class using k-means clustering \cite{arthur2007k}.
For each of the classes or discrete values in the data set, k-means clustering is applied to find $n$ clusters, where $n$ is the number of remaining samples in the reduced data set.
After the clustering step, that is, when $n$ central points in the input space have been found, according to the surrounding data points, the closest sample in the data set is found an used as a representative data sample.
Finally, the reduced data set consists of the representative data samples for all $n$ clusters.
By finding the closest sample to the central point, it is guaranteed that the reduced training set only consists of samples present in the original data set.

Similar selection strategies are used to reduce training set sizes in instance-based machine learning algorithms \cite{Wilson2000}, such as k-nearest neighbour methods.
We have experimented with the application of these strategies, from \cite{Chen1996,Sanchez2004}, but found the achieved reduction to be small and not effective. This effect might be caused by the high dimensionality of the considered data set and should be investigated as part of future work.

\subsubsection{Loss-based Subset Selection}

The loss, also cost function, describes how well the DL model approximates the desired output for a given output and is important information to calculate the gradients for backpropagation errors throughout the network, i.e. the actual training step.
A higher loss indicates an input sample for which the model output largely differs from the expected output, where as a low loss indicates conformity with the expected output.

The loss information can therefore be used to identify the most difficult respectively least difficult samples in the data set in respect to the model's current state.

Loss-based subset selection does exploit this basic idea by recording the initial loss for all samples in the data set for an untrained model after it has been initialized.
While this initialization is performed randomly, many of the complex DNN architectures follow specifically designed initialization strategies \cite{Bengio2012a} to control the range of the initial model parameters and to improve model robustness.
By repeatedly measuring the initial loss of training samples over different model initializations, it is therefore possible to identify samples which are more difficult for the model, i.e. they result in a higher initial loss, than other samples.

\begin{figure}
    \centering
    \input{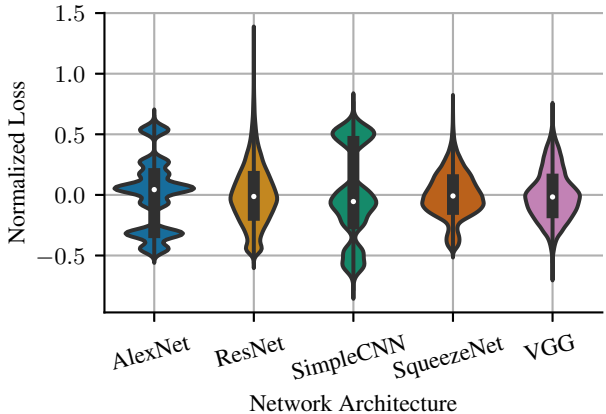}
    \caption{Distribution of initial losses for the images in the CIFAR-10 data set on five model architectures. The loss values have been recorded over 10 random seeds and are normalized for comparability.}
    \label{fig:initiallosses}
\end{figure}

To test this hypothesis and to understand the actual behaviour of DL systems, we have measured the initial losses of all training samples in the CIFAR-10 data set for five model architectures, namely AlexNet \cite{Krizhevsky2014a}, ResNet-18 and Simple CNN (as described in Section~\ref{sec:architectures}), SqueezeNet \cite{Iandola2016}, and VGG-16 \cite{Simonyan2015}.
The distributions of the initial loss values are show in \figurename~\ref{fig:initiallosses}. 
The loss per image is averaged over 10 runs with different random seeds.

Although different random model initialization are considered, a distribution of sample losses can be observed.
Both for AlexNet and SimpleCNN, there are groups of samples, that have especially low or high losses and are respectively easy or difficult for the initial model.
For ResNet, the distribution is more closely to a normal distributions, with the largest group of samples being unaffected by random model initializations, similar to the other models, but there is also a number of outliers with higher average loss.
These outliers also indicate a difference in difficulty for smaller subset of samples.

\subsubsection{Random Subset Selection}

Random selection of a subset of instances from the full data set is the baseline approach and usually the current best practice, when only using a subset of the data to test a model implementation.
In general, if the size of the random subset is large enough or the selection is repeated multiple times, random sampling should result in a representative selection of training samples.
However, for TSR the interest is to find small subsets, if possible, as the size of the effective training set is correlated to the duration of model training.

Nevertheless, random subset selection is the easiest approach to reduce the size and complexity of an initially large original data set and does not require additional computation, expect sampling random numbers.
Therefore it is the current best practice \cite{Dwarakanath2018,karpathy2016cs231n}, especially when testing initial model developments.

\subsection{Limitations} \label{sec:limitations}

TSR decreases the number of training samples and thereby the amount of available data for training the model.
Due to the data-driven nature of ML systems, decreased data availability will prohibit the trained model from being as expressive and well-performing as the model trained with the original full data set.
This is an inherent limitation of TSR to be noted.
However, it is also not the goal of TSR to produce high-quality models with less data, and thereby improve the training process, but the provide a technique for efficient model testing and to produce models, that can serve as proxies during testing.

Another aspect to consider for the application of TSR is \textit{overfitting}, i.e. a ML model fully captures the structure and information in the training, thereby achieves high performance on the training data, but does not generalize on the validation data set.
Overfitting can occur if the model capacity, e.g. the number layers and nodes in a neural network, is too large in comparison to the complexity and size the training data set.
In the following experiments, we will observe overfitting, but to different extents in respect to the reduced training set size and the corresponding TSR strategy.

\section{Evaluation} \label{sec:evaluation}

\subsection{Experimental Setup} \label{sec:expsetup}

We evaluate the previously presented training set reduction strategies on an image classification case study with 10 classes.
As a dataset, we use the CIFAR-10 dataset \cite{Krizhevsky2014cifar}, consisting of training and validation with 50000 and 10000 labelled images.
All experiments are performed on two model architectures, a commonly, well-performing ResNet-18 model and a simple CNN with two convolutional layers and one fully-connected dense layer.
Both model implementations are taken from the PyTorch model zoo respectively documentation materials.
We perform the experiments with two model architectures to understand differences in the effect of training set reduction for large and small architectures.
Furthermore, the SimpleCNN model has little capacity whereas the ResNet model is also applicable to larger data sets than CIFAR-10. This allows to evaluate effects of under- and overfitting on the data sets.

\begin{figure*}[t]
    \centering
    \input{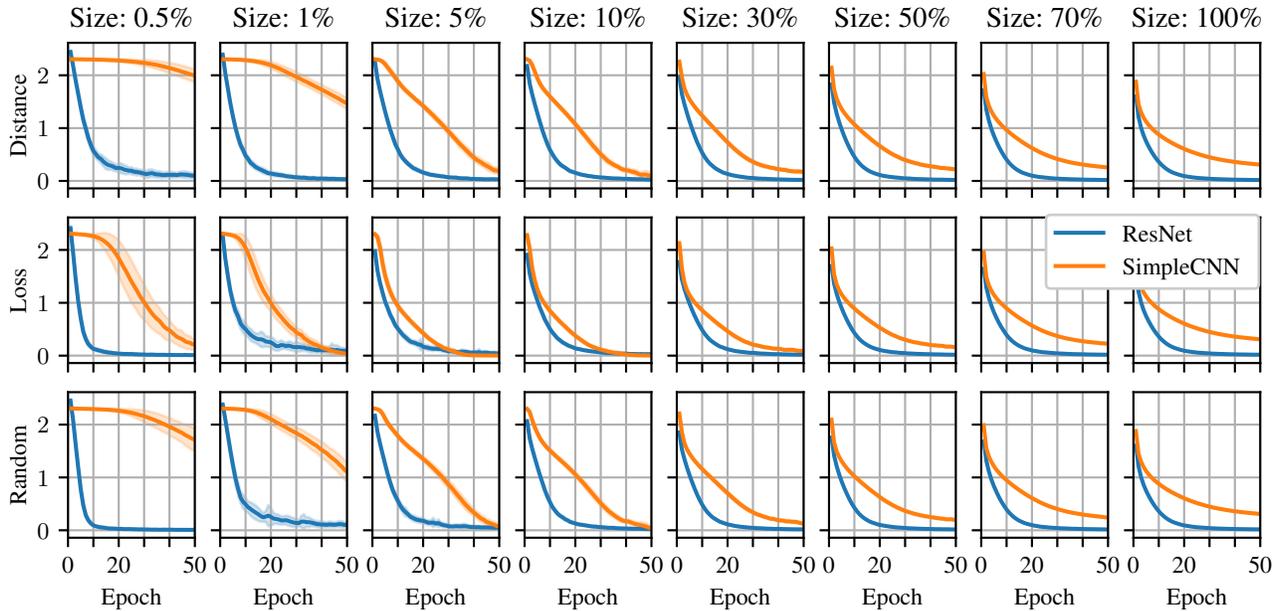}
    \caption{Training loss curves for reference runs with different reduction techniques and training and validation set sizes. The curves for the full size are identical for all three techniques.}
    \label{fig:refruns_train}
\end{figure*}

\subsection{Results}

\begin{figure*}[t]
    \centering
    \input{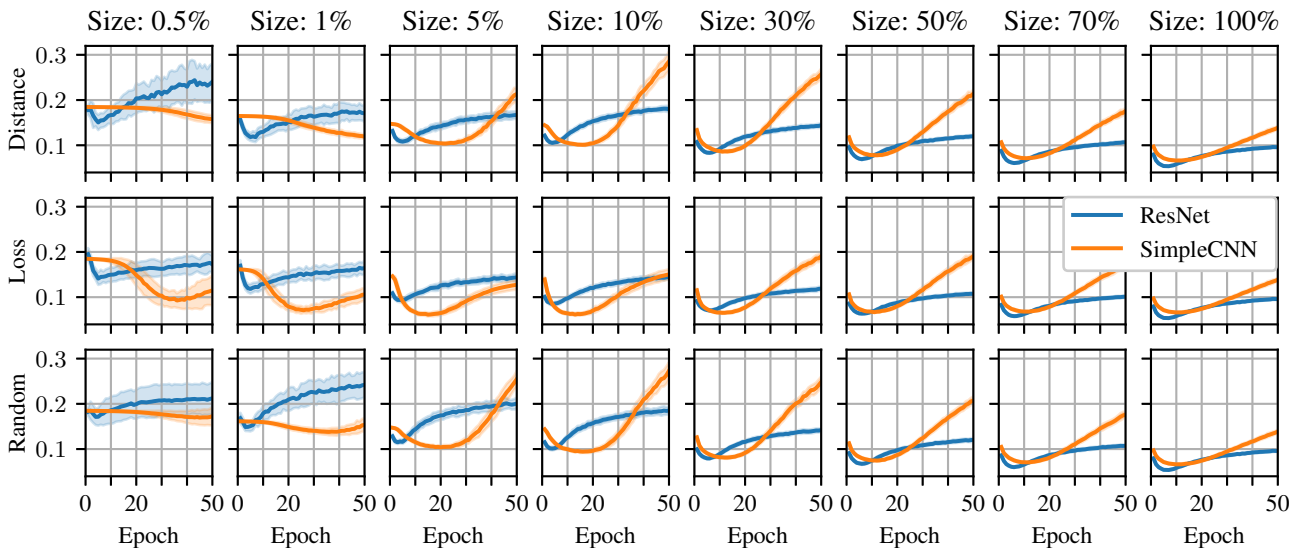}
    \caption{Validation loss curves for reference runs with different reduction techniques and training and validation set sizes. The curves for the full size are identical for all three techniques.}
    \label{fig:refruns_val}
\end{figure*}

The averaged loss curves for training the SUTs with all combinations of TSR strategy and resulting data set size are shown in \figurename~\ref{fig:refruns_train} for training and \figurename~\ref{fig:refruns_val}.
The rightmost column, which is identical in all three rows, shows the training loss curves when using the original full-sized data set.
We will refer to these curves as the full-sized behaviour.

Generally, as it is expected, we observe a closer approximation of the full-sized curves by an increasing data set size for all three TSR strategies.
While especially the loss curves for the SimpleCNN model diverge from the reference negative exponential curve, an increase to at least 10\% of the original data set size shows a general resemblance of the final loss curve.
However, even visually, the degree of similarity varies between the three strategies for the different data set sizes.

For a more precise evaluation of the differences than visual expectation, the results are next compared numerically.
For each combination of TSR strategy and reduced data set size, the difference between the loss curves is approximated by the similarity in parameters describing a fitted curve to the measured data.

When training DL systems, the shape of the loss curve is an indicator for the appropriate parametrization of the model. The loss should be exponentially decreasing, resembling a “hockey-stick shape" when plotted (see \cite{karpathy2016cs231n} for a practical discussion of different loss curve shapes).
The training loss curve is therefore modeled as an exponential function with a negative exponential coefficient: $y = ae^{-bx} + c$.
For full-sized model training, the validation loss curve should express a similar form to the training curve.
However, as a result of both the reduced data set size, as discussed before in Section~\ref{sec:limitations}, the network does not achieve similar performance, and therefore the test loss curves do not follow the previous assumptions. 
Instead, we choose a fifth-order polynomial to model the test loss, which more closely resembles the actual curves.

The similarity metric is than computed as the mean squared error between fitted curve parameters for the reduced training losses and the full-sized training losses.
While this metric does not give a specific dimension to compare the different curves on, it does serve as a relative metric, where a lower error indicates higher similarity.

The similarity between loss curves for reduced data set sizes and the full-sized data set is further depicted in Table~\ref{tab:curve_distances}.
First, the results confirm the previous visual results and the general intuition of the model's behaviour by showing higher similarity for larger data subsets in all columns.
Second, the results further show, that an aggressive reduction to less than 5\% of the original size does result in very different behaviour for the larger ResNet model.
For SimpleCNN, the smaller model architecture, the subset size has an even stronger influence, here the similarity drastically reduces with less than 10\% of the original size.

Considering the performance differences between the three TSR strategies, we observe best performance for the loss-based subset selection.
Loss-based subset selection decides which training samples to include based on the initial loss of these samples on untrained instances of the ML model.
This strategy has the closest connection to the actual behaviour, that is to be imitated.

The other two strategies, distance-based and random subset selection, can also approximate the full-sized training behaviour, especially with larger subset sizes.
Selecting training samples based on their distance with the goal to evenly cover the input space does not show to be effective for imitating the training behaviour.
This strategy performs even slightly worse than random subset selection, while requiring higher initial efforts for clustering the full-sized data set.

However, loss-based selection achieves similar characteristics, when measured on the curve similarity, than the other strategies with less training data, or, respectively, better approximation with the same amount of training data.
While loss-based subset selection requires the initial preparation step of gathering loss values, this step can be achieved with comparatively low cost as it requires forward processing of the ML model without backpropagation of the errors.
This backpropagation step, the actual training step, is much more costly, while the forward processing is the same step as querying the ML model for a prediction for a new input sample.

\begin{table}
    \centering
    \subfloat[Training Loss Curves]{\begin{tabular}{lrrr|rrr}
\toprule
Size & \multicolumn{3}{c}{ResNet} & \multicolumn{3}{c}{SimpleCNN} \\
(in \%) & Distance &   Loss & Random &  Dist. &   Loss & Rand. \\
\midrule
0.5   &    1.121 &  1.694 &  1.845 &       --- &    --- &    --- \\
1     &    1.112 &  0.777 &  1.096 &       --- &  2.196 &    --- \\
5     &    0.870 &  0.408 &  0.769 &     5.967 &  1.221 &  7.064 \\
10    &    0.760 &  0.371 &  0.655 &     1.727 &  0.929 &  1.766 \\
30    &    0.503 &  0.232 &  0.371 &     0.796 &  0.481 &  0.707 \\
50    &    0.308 &  0.144 &  0.232 &     0.443 &  0.298 &  0.400 \\
70    &    0.163 &  0.073 &  0.122 &     0.222 &  0.162 &  0.202 \\
\bottomrule
\end{tabular}
}\\
    \subfloat[Validation Loss Curves]{\begin{tabular}{lrrr|rrr}
\toprule
Size & \multicolumn{3}{c}{ResNet} & \multicolumn{3}{c}{SimpleCNN} \\
(in \%) & Distance &   Loss & Random &  Dist. &   Loss & Rand. \\
\midrule
0.5   &    0.107 &  0.121 &  0.102 &     0.084 &  0.084 &  0.085 \\
1     &    0.077 &  0.086 &  0.082 &     0.065 &  0.052 &  0.060 \\
5     &    0.047 &  0.024 &  0.040 &     0.054 &  0.068 &  0.057 \\
10    &    0.038 &  0.017 &  0.029 &     0.050 &  0.046 &  0.049 \\
30    &    0.029 &  0.012 &  0.021 &     0.033 &  0.015 &  0.024 \\
50    &    0.018 &  0.008 &  0.012 &     0.020 &  0.007 &  0.015 \\
70    &    0.009 &  0.004 &  0.007 &     0.010 &  0.003 &  0.007 \\
\bottomrule
\end{tabular}
}
    \caption{Similarity of loss curves for reduced and full data sets. A smaller value indicates a higher similarity between curves (---: unusable data, curve fitting failed).}
    \label{tab:curve_distances}
\end{table}

\subsection{Threats to Validity}
\subsubsection{Internal}
The first threat to internal validity is the influence of stochastic aspects within the experiments.
Many parts of training ML models, especially neural networks, include stochastic elements, albeit the order of input samples during training or model initialization.
Our TSR strategies also rely on stochastic decisions.
Distance-based subset selection uses k-means clustering, which randomly initializes the clusters at the beginning, loss-based subset selection depends on model initializations, and random subset selection is purely random.
To mitigate this risk, we repeated all experiments 40 times and report average results.
We further use a k-means clustering method that takes several cluster initializations and internally decides on the best result.

The influence of hyperparameter choices has been reduced by keeping the same set of parameters throughout all experiments. This affects especially the model training. Adjusting the training hyperparameters might allow to get closer approximation of the desired training behaviour, however, it is not desired to adjust central parameters - that are part of the model tuning - for testing purposes. Instead it is necessary to also test the hyperparameters.

Finally, we try to avoid faults in our implementation by relying on stable external libraries and implementations where possible (see Section~\ref{sec:expsetup}).

\subsubsection{External}

Our evaluation uses an openly available and widely known data set, the CIFAR-10 data set \cite{Krizhevsky2014cifar}.
We further rely on evaluation systems, i.e. the network architectures, that are either basic approaches for the task, such as SimpleCNN, or resemble state-of-the-art modeling aspects for deep neural networks, such as ResNet.

\subsubsection{Construction}

A threat to construction is the assumption, that a similarity in loss curves can be a useful indicator for testing the underlying ML model.
While the loss curve is most likely not the only useful indicator and others should be considered as well, it is a metric that is used by both practitioners \cite{karpathy2016cs231n} and researchers \cite{Dwarakanath2018}.
To further address this threat, we propose to consider other metrics and similarity functions in future work.

\section{Conclusion}
We have discussed and evaluated three training set reduction (TSR) strategies.
The goal of TSR is to find a smaller subset of the original training data set, that shows similar characteristics during model training.
Potential use cases for TSR are faster model training for software testing, e.g. in continuous integration environments.

Of the three considered strategies, we found both systematic approaches, i.e. distance-based reduction, based on clustering the training samples, and loss-based reduction, based on selecting instances with high initial loss values, to be more effective than random subset selection.
We performed an experiment with two deep neural network architectures, a simple CNN and a state-of-the-art ResNet, on the CIFAR-10 data set with 50,000 training samples.
Data sets with only 10-50\% of the initial size closely resembled the loss curves of the full-sized training process.

Our initial study showed systematic TSR to be a promising approach for data set reduction and we encourage the usage when selecting a representative subset for frequent testing purposes.

For future work, we plan to explore the effectiveness of TSR for testing learning systems on other domains and experimental settings. We further want to explore the appropriateness of the reduced data subset for model-tuning aspects, where both reducing the runtime of the system and showing similar behaviour to the full-sized model are important.

\section*{Acknowledgements}
The experiments were performed on the Abel Cluster, owned by the University of Oslo and Uninett/Sigma2, and operated by the Department for Research Computing at USIT, the University of Oslo IT-department. \url{http://www.hpc.uio.no/}

\printbibliography
\end{document}